\documentclass[runningheads]{llncs}
\usepackage[T1]{fontenc}
\usepackage{graphicx}
\usepackage{amsmath}
\usepackage{amssymb}
\usepackage{array}
\usepackage{booktabs}
\usepackage{pifont}
\usepackage{bm}
\usepackage{url}
\usepackage{hyperref}
\hypersetup{
  hidelinks,
  bookmarksnumbered=true
}
\newcommand{\cmark}{\ding{51}}
\newcommand{\xmark}{\ding{55}}
\newcommand{\corrauth}{\textsuperscript{\ensuremath{\dagger}}}
\begin{document}
\title{FedVideoMAE: Efficient Federated Video\texorpdfstring{\\}{ }Moderation with Differential Privacy\texorpdfstring{\\}{ }and Secure Aggregation}
\titlerunning{FedVideoMAE: Federated Video Moderation with DP and SA}
\author{Ziyuan Tao\inst{1,2}\corrauth \and
Chuanzhi Xu\inst{2}\corrauth \and
Sandaru Jayawardana\inst{2} \and
Adnan Mahmood\inst{1} \and
Wei Bao\inst{2} \and
Kanchana Thilakarathna\inst{2} \and
Teng Joon Lim\inst{3}}
\authorrunning{Tao et al.}
\institute{\fontsize{9}{11}\selectfont
School of Computing, Macquarie University, Sydney, Australia
\and
School of Computer Science, The University of Sydney, Sydney, Australia
\and
School of Electrical and Computer Engineering, The University of Sydney, Sydney, Australia
\\[0.55em]
\ensuremath{\dagger}
\email{ziyuan.tao@students.mq.edu.au,} \email{chuanzhi.xu@sydney.edu.au}
}
\maketitle
\begin{abstract}
\begingroup\sloppy
Short-form video moderation is increasingly pushed toward edge and privacy-sensitive settings, where users may intend videos for a limited audience, such as friends or private groups, but sending raw clips to a central server can broaden exposure, consume bandwidth, and add moderation latency. Federated learning can keep videos on device, but unprotected model updates may still leak information, and full-video backbones are expensive to communicate. We present \emph{FedVideoMAE}, a privacy-preserving federated framework for violence detection that adapts a frozen VideoMAE backbone with lightweight LoRA and prompt parameters. Each training round combines self-supervised masked video reconstruction with client-side differential privacy and pairwise masked aggregation (SA) of adapter updates. Violence labels are held out from federation and used only for downstream evaluation, separating private representation learning from supervised assessment. On RWF-2000, exchanging 5,518,848 trainable parameters instead of the 156,371,328-parameter instantiated pretraining state gives a $28.3\times$ model-state payload ratio. FedVideoMAE reaches 77.25\% test accuracy without DP or SA, while accuracy under DP+SA remains in the 65.25--66.00\% range. Transfer experiments on RLVS and binary UCF-Crime show similar behavior. These results characterize the privacy--utility trade-off for edge video moderation. Code is available at: \url{https://github.com/zyt-599/FedVideoMAE}
\endgroup

\end{abstract}
\section{Introduction}

Short-form video platforms are now a major channel for online communication~\cite{violot2024shorts}, with billions of daily uploads across major services. This scale makes content moderation both more important and harder to deploy. At this scale, centralized cloud pipelines combine perceptual hashing, multimodal deep models~\cite{tsai2021multimodal}, and human review. Recent industrial systems further use MLLM-based cascade designs to improve moderation coverage while reducing the cost of full-scale model inference~\cite{wang2025filter}. However, in on-device or privacy-sensitive settings, moderation quality must be balanced against three deployment constraints.

First, transmitting raw or lightly processed clips to cloud servers creates privacy exposure, especially for restricted-visibility content such as private albums, friend-only posts, or sensitive personal recordings. Recent studies show that uploaded or unpublished videos, and data reused for model training, can still leak sensitive information~\cite{Luo_2024_WACV,do2025vbda}. Second, short-form video is bandwidth-intensive~\cite{gong2025cetc,do2025vbda}: platforms may ingest large files later discarded for policy reasons, incurring substantial wasted bandwidth and cost. Third, remote inference adds moderation latency~\cite{gong2025cetc,xu2025edgedl} through upload-and-infer round-trips that are undesirable in interactive or near-real-time use. Although benchmarks such as RWF-2000~\cite{cheng2021rwf2000,negre2024violenceReview,sultani2018ucfcrime} have advanced violence detection, most deployment paths still inherit these cloud-centric privacy and communication burdens.

\begin{figure}[!t]
\centering
\includegraphics[width=\textwidth]{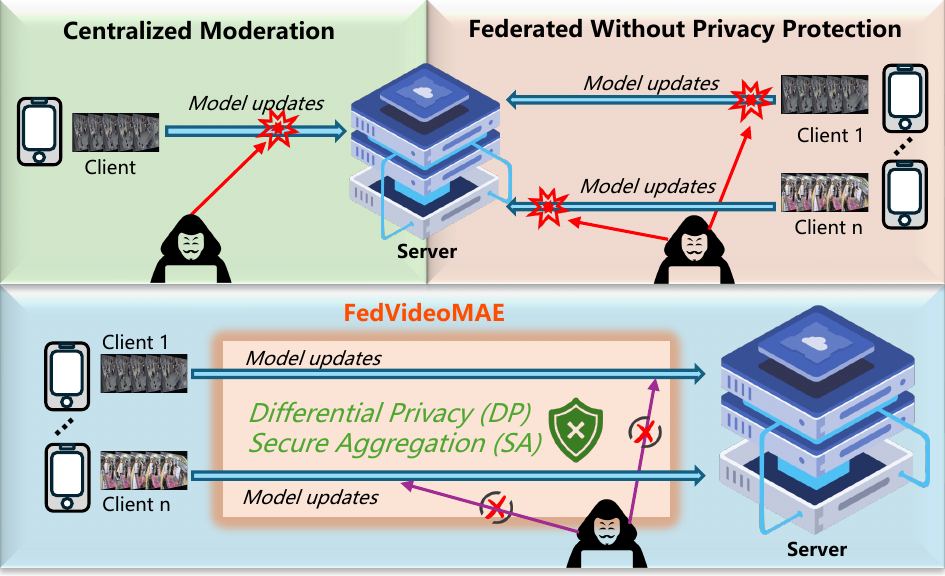}
\caption{Three deployment paradigms for video moderation: centralized cloud upload (exposure in transit or on the server), basic federated learning with unprotected updates (gradient leakage), and our federated pipeline with client-side DP-SGD and pairwise masking of adapter updates.}
\label{fig:with_and_without_privacy}
\vspace{-0.8em}
\end{figure}

Federated learning (FL)~\cite{mcmahan2017fedavg,saha2024privacyflsurvey} fits this setting because private videos can remain on device while clients train a shared model. Figure~\ref{fig:with_and_without_privacy} also shows why standard FL is incomplete: raw videos stay local, but unprotected model updates can reveal content through gradient inversion and related attacks~\cite{wu2023simple,e2egi2023,du2024sok}. This calls for layered protection that combines formal privacy with protocol-level hiding. Existing FL methods for video understanding still leave three problems unresolved: (i)~high communication cost from large spatiotemporal transformers, (ii)~limited integration of differential privacy (DP) and secure aggregation (SA), and (iii)~little analysis of how DP noise interacts with scarce per-client data and parameter-efficient fine-tuning (PEFT).

We propose FedVideoMAE for resource-constrained video moderation, combining DP-SGD with pairwise masked aggregation (Fig.~\ref{fig:with_and_without_privacy}, bottom). FedVideoMAE keeps a frozen self-supervised VideoMAE backbone on device and federates only LoRA adapters and reconstructive prompts through masked video reconstruction, while violence labels are reserved for a later centralized evaluation stage. Client-side DP-SGD clips per-sample adapter gradients and adds Gaussian noise, with the reported $\varepsilon$ values given as RDP-accountant estimates under the stated sampling approximation. The server then aggregates pairwise masked adapter updates, so raw videos and intermediate features remain local while federation communicates only adapter states.

FedVideoMAE is designed for a difficult video-FL setting in which each device holds few clips, the pretrained model is large, and DP-SGD acts on a compact trainable subspace. In our 40-client RWF-2000 setup, 5,518,848 of 156,371,328 instantiated parameters (3.5\%) are federated, giving a $28.3\times$ total-state-to-uploaded-state ratio. Small-head test accuracy is 77.25\% without DP or SA and 65--66\% under the reported DP configurations, with ROC-AUC above 71\% under DP+SA. We compare these results with full-model FedAvg and FedProx references on the same partition and with $\varepsilon=1$ transfer on RLVS and binary UCF-Crime. An effective-SNR heuristic organizes the influence of local data scale, DP noise, trainable-parameter fraction, and participation on adapter optimization. Therefore, our main contributions are:
\begin{itemize}
    \item \textbf{FedVideoMAE}: an on-device federated MAE+PEFT pipeline with integrated DP-SGD and SA. The method separates federated self-supervised representation learning from centralized supervised moderation evaluation, keeps raw videos local, and communicates only compact adapter states.
    \item \textbf{Privacy--utility evidence for edge moderation}: a controlled 40-client RWF-2000 study showing how much accuracy is retained when compact adapter updates are protected by DP+SA, together with full-model FL references and auxiliary transfer checks that place the trade-off in context.
    \item \textbf{Effective-SNR heuristic}: an optimization-side diagnostic linking client data scale, DP noise, trainable-parameter fraction, and participation. 
\end{itemize}

\section{Related Work}

\textbf{Federated Video Learning.}
Recent work has extended federated learning to video understanding. Rehman et al.~\cite{rehman2022fedvssl} propose FedVSSL for self-supervised video representation learning, showing that useful spatiotemporal features can be learned without centralizing raw clips. Yang et al.~\cite{yan2023fedmae} introduce FedMAE with one-block masked autoencoders to reduce communication by $20$--$30\times$ and show that video FL is more demanding than image FL: the models are larger, the inputs are temporally redundant but high-dimensional, and client distributions can differ sharply in scene dynamics and motion statistics. Their work focuses on representation quality and communication reduction, with less emphasis on formal privacy guarantees or privacy noise under small per-client moderation datasets. FedVideoMAE targets violence-oriented moderation under a two-stage protocol (federated MAE pretraining, centralized supervised evaluation) with integrated DP-SGD and SA on adapter uploads.

\textbf{Parameter-Efficient Fine-Tuning.}
He and Wang~\cite{he2024pmae} propose pMAE for federated continual learning with reconstructive prompts and LoRA~\cite{hu2022lora}, reducing communication overhead. LoRA preserves pretrained weights, avoids extra inference branches, and keeps the trainable set small, which fits video transformers dominated by large linear layers. In federated settings, this matters because uplink cost scales with the trainable state that clients exchange, not with the frozen backbone kept locally. FedVideoMAE adopts a similar PEFT design on VideoMAE-Base and couples masked reconstruction with privacy tools for edge moderation. This coupling also changes the privacy--utility problem: DP noise is injected into a compact adapter subspace whose optimization behavior may differ from full-model fine-tuning. Existing PEFT federated methods do not jointly study DP and SA on VideoMAE-scale models with scarce per-client data.

\textbf{Differential Privacy in Federated Settings.}
Differentially private federated learning is well studied~\cite{fu2026dpflreview}. DP-SGD~\cite{abadi2016dpsgd} applies per-sample gradient clipping and Gaussian noise, and Rényi differential privacy~\cite{mironov2017rdp} gives tighter composition accounting over many rounds. Much of this literature uses image or language benchmarks with more local data per client than realistic on-device video moderation. Recent work has explored optimization and aggregation strategies to reduce DP-induced utility loss~\cite{fu2026dpflreview,bienstock2025dmm,ball2024securestateful}, but these results mostly come from larger-scale settings and do not directly cover PEFT-based video transformers with very limited local shards. For video, Luo et al.~\cite{Luo_2024_WACV} show that DP-SGD can substantially degrade activity-recognition accuracy, illustrating the difficulty of applying DP to high-dimensional spatiotemporal models. The problem becomes sharper in our setting, where privacy noise acts on a small trainable subspace and each client holds only tens of clips. With so few clips per client, the sampling rate and repeated-round composition make the privacy budget a direct optimization constraint alongside its accounting role. Accordingly, we report sample-level, per-client privacy under the stated client-wise accounting design.

\textbf{Gradient Leakage and Layered Protection.}
Federated learning keeps raw data local, yet shared updates can reveal training examples under adaptive attacks~\cite{wu2023simple,e2egi2023,du2024sok}. These findings motivate combining local retention with formal privacy and protocol-level protection. Secure aggregation prevents an honest-but-curious server from inspecting individual client updates~\cite{bonawitz2017secagg}. Differential privacy bounds what can be inferred from released aggregates or the final model~\cite{dwork2014dp,abadi2016dpsgd}. In the experimental system, pairwise masks are applied to adapter states before averaging, while DP-SGD limits inference under the stated sample-level accounting. Compact adapter updates are cheaper to transmit, but they can still encode example-specific directions. Most prior studies evaluate these tools on image or language models. Their behavior in parameter-efficient video transformers with small client datasets remains less studied, motivating our framework and ablations.

\begin{figure}[!ht]
\centering
\includegraphics[width=\textwidth]{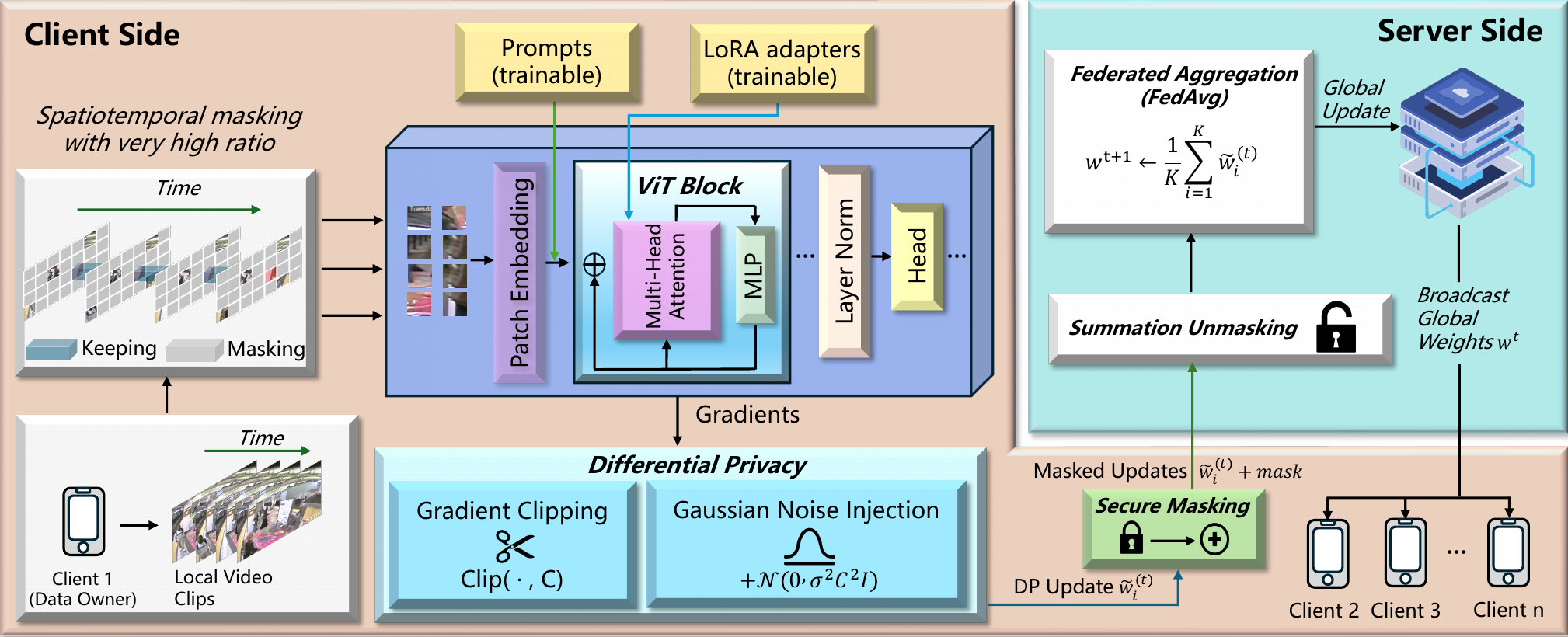}
\caption{High-level architecture of FedVideoMAE. Clients perform federated masked-video reconstruction on frozen VideoMAE features with LoRA and reconstructive prompts, while DP-SGD and pairwise masking are applied to adapter states before server aggregation. Violence classification is evaluated in a separate centralized stage.}
\label{fig:architecture}
\vspace{-1.2em}
\end{figure}

\section{Methodology}

\subsection{Federated Training Pipeline}

Figure~\ref{fig:architecture} summarizes FedVideoMAE as a round-based pipeline in which the server broadcasts global LoRA and prompt parameters to multiple clients and aggregates their updates with FedAvg. On each client, the base weights of the pretrained VideoMAE encoder and decoder~\cite{tong2022videomae} remain frozen, while LoRA adapters in both transformer stacks and a reconstructive prompt are trainable. Local clips pass through the frozen base network, and the adapted decoder reconstructs held-out spatiotemporal patch tokens using a tube-masking ratio of 0.9 and normalized masked-pixel targets. The reconstructive prompt provides a compact adaptation path alongside LoRA injections in attention and MLP blocks. Federation optimizes masked-reconstruction MSE with light $\ell_2$ regularization only, so no violence classifier or violence label is used on device. Figure~\ref{fig:architecture} shows the full moderation pipeline, with the classifier path reserved for centralized evaluation. Only LoRA and prompt states are optimized and communicated, while raw videos, intermediate features, and frozen base weights remain on device.

Each client applies DP-SGD to adapter gradients before upload, followed by pairwise masking and server-side FedAvg aggregation (Sec.~\ref{sec:privacy}). After federated MAE adaptation, we freeze the encoder and evaluate violence detection through centralized linear probing and 20-epoch small-head fine-tuning, preserving the separation between federated representation learning and supervised moderation. The complete pipeline comprises (1)~on-device federated MAE with PEFT, (2)~DP-SGD and pairwise masking of adapter uploads, (3)~server-side FedAvg, and (4)~centralized probe and head evaluation on RWF-2000 labels.

\subsection{Privacy Mechanisms in FedVideoMAE}
\label{sec:privacy}

\textbf{Differential Privacy via DP-SGD.}
We adopt DP-SGD~\cite{abadi2016dpsgd} with Opacus~\cite{opacus} and RDP accounting~\cite{mironov2017rdp}. Privacy is enforced at the sample level on each client's local clips: in every minibatch, per-sample gradients are clipped to $\ell_2$ norm $C$, summed, and perturbed with Gaussian noise before averaging:
\begin{equation}
\bar{g}_t(x_i) = \frac{g_t(x_i)}{\max\left(1,\frac{\|g_t(x_i)\|_2}{C}\right)},\quad
\tilde{g}_t = \frac{1}{|B_t|}\!\left(\sum_{x_i \in B_t}\bar{g}_t(x_i) + \mathcal{N}(0,\sigma^2 C^2 I)\right).
\end{equation}
For each participating client, the RDP accountant estimates sample-level privacy loss on its local shard $D_i$ using its Poisson-sampling approximation for fixed-size minibatches. The estimate applies to samples within a local shard and does not provide client-level privacy or a joint guarantee for the final global model. With $q_i=\min(1,|B|/|D_i|)$, $E$ local epochs per round, and $R$ rounds, we compose $T_i=\lceil|D_i|/|B|\rceil\cdot E\cdot R$ DP steps when the client participates every round. For RWF-2000 federated MAE ($|B|=32$, $|D_i|=34$ train clips per client, $E=4$, $R=200$), the schedule yields $q_i=32/34$ and $T_i=1600$. We fix $C=1.0$ and $\delta=10^{-5}$, use fixed-size minibatches with Poisson subsampling disabled, and calibrate $\sigma_i$ with Opacus \texttt{get\_noise\_multiplier} over $T_i$ steps. Table~\ref{tab:dp_accounting} therefore reports RDP-accountant estimates under this approximation. Opacus hooks add noise only to trainable adapter and prompt gradients during the client backward pass, while frozen encoder and decoder weights are excluded from both optimization and perturbation. Applying noise only to trainable coordinates leaves clipping sensitivity controlled by $C$ and changes both the total perturbation norm and its interaction with adapter optimization.

\textbf{Secure Aggregation.}
The experimental implementation simulates pairwise masked aggregation of adapter states after local DP-SGD, following the mask-cancellation principle of secure aggregation~\cite{bonawitz2017secagg}. With full participation, pairwise masks cancel in the aggregate, yielding the averaged adapter state.

\textbf{Threat Model.}
The evaluated setting keeps raw videos and client features on device and applies local DP-SGD with simulated pairwise masked aggregation to adapter states. The privacy analysis covers sample-level accounting and mask cancellation under full participation, while Byzantine behavior, poisoning, backdoor attacks, and endpoint side channels remain outside this scope.

\subsection{LoRA-Based Adaptation and Communication Cost}

\textbf{LoRA Formulation.}
FedVideoMAE freezes the VideoMAE-Base weights and trains LoRA adapters~\cite{hu2022lora} together with reconstructive prompts. For pretrained weight $W_0$, LoRA constrains the update to rank-$r$ factors:
\begin{equation}
\Delta W = \frac{\alpha}{r}BA,\qquad
A \in \mathbb{R}^{r \times k},\;
B \in \mathbb{R}^{d \times r},\;
r \ll \min(d,k),
\label{eq:lora}
\end{equation}
with only $A$ and $B$ optimized. We use $r=16$ and $\alpha=32$. The implementation matches five linear layers in each transformer block: \texttt{q\_proj}, \texttt{k\_proj}, \texttt{v\_proj}, \texttt{fc1}, and \texttt{fc2}. Across 12 encoder blocks and 8 decoder blocks, this gives $12\times5+8\times5=100$ LoRA modules and 3,145,728 LoRA parameters. An 8-token reconstructive prompt of dimension 1536 and its projection contribute another 2,373,120 parameters, yielding exactly 5,518,848 federated trainable parameters. The instantiated pretraining model contains 156,371,328 parameters in total, so the federated state is approximately 3.5\% of the instantiated model. Encoder and decoder base weights remain frozen, while their LoRA adapters and the reconstructive prompt are optimized during federated MAE. The downstream MLP head is trained centrally after federation and is excluded from the uplink.

\textbf{Communication Cost.}
Each round exchanges the 5,518,848 LoRA and prompt parameters instead of a 156,371,328-parameter instantiated pretraining state, giving a total-state-to-uploaded-state ratio of approximately $28.3\times$. Relative to the 86.2M-parameter supervised updates used by the FedAvg and FedProx references, the same payload is approximately $15.6\times$ smaller. Both ratios measure transmitted model states, and these parameter-count comparisons exclude coordination traffic and protocol metadata.

\subsection{Effective-SNR Heuristic}
\label{sec:snr}

We use an effective-SNR heuristic to organize optimization-side factors in the observed privacy--utility behavior, separately from the formal DP accounting above. For a client with $n$ samples, we compare its averaged gradient signal with DP-SGD noise of scale $\sigma C$:
\begin{equation}
\mathrm{SNR}
    = \frac{\|\mathbb{E}[\bar{g}_t]\|_2}{\sigma C}
    \propto \frac{\sqrt{n}}{\sigma}.
\label{eq:snr}
\end{equation}
Clipping sensitivity is controlled by $C$, while trainable-state dimension affects the perturbation norm and geometry. We summarize PEFT scale and participation relative to a chosen reference as:
\begin{equation}
\mathrm{SNR}_{\mathrm{eff}}
    = \mathrm{SNR}
      \cdot \sqrt{\frac{P_{\mathrm{train}}}{P_{\mathrm{total}}}}
      \cdot \sqrt{\frac{N_{\mathrm{client}}}{N_{\mathrm{ref}}}},
\label{eq:snr_eff}
\end{equation}
where $N_{\mathrm{ref}}$ denotes the fixed comparison cohort. The square-root factors normalize trainable-state size and client participation relative to the reference configuration. RWF-2000 uses $P_{\mathrm{train}}/P_{\mathrm{total}}=0.035$ and $N_{\mathrm{client}}=40$ in every round. Since $\sigma$ decreases from 152.5 to 35.9 between $\varepsilon=1$ and $5$, the identical 65.25\% accuracies indicate an empirical plateau shaped by the full optimization setting.

\section{Experiments}
\label{sec:experiments}

\subsection{Experimental Setup}

\textbf{Datasets.}
Our main federated study uses RWF-2000~\cite{cheng2021rwf2000}, a surveillance violence benchmark with 2,000 videos (1,000 violent and 1,000 non-violent). We adopt the standard split with 1,600 development clips and 400 test clips (1,360 train, 240 validation, and 400 test clips). The 1,360 training clips and 240 validation clips are each globally shuffled and assigned round-robin to 40 clients, giving each client 34 train clips for federated MAE and 6 validation clips for supervised FedAvg and FedProx. Federated MAE and DP-SGD use $|D_i|=34$ clips per client, and no separate client-side validation split is held out during federation. Across each client's combined 40-clip development shard, comprising 34 training and 6 validation clips, the labels remain heterogeneous because assignment is not stratified by class. Each shard contains $20.0 \pm 13.4$ violent clips on average, corresponding to a positive ratio of $0.50 \pm 0.34$, with a range of 0--40 violent clips and 5/40 single-class shards. We describe the split as approximately label-balanced at the corpus level but not strictly IID across clients. The same 40-client partition is used for FedVideoMAE and full-model FedAvg and FedProx references, matching the small local datasets expected in on-device moderation.

Beyond RWF-2000, we evaluate $\varepsilon=1$ transfer on RLVS~\cite{soliman2019rlvs} and binary UCF-Crime~\cite{sultani2018ucfcrime}. RLVS is a short-video violent and non-violent dataset (200 held-out clips) and is close to our main task. UCF-Crime was originally proposed for long-form anomaly detection, and we use a simplified normal-vs-abnormal binary setup with 440 held-out clips. This transfer case is harder because the videos are longer and ``abnormal'' covers more than violence.

\textbf{Federated Pretraining Configuration.}
All experiments share the same federated setup unless noted otherwise: 200 communication rounds with all 40 clients participating each round, 4 local epochs per round, batch size 32, AdamW with learning rate $2\times 10^{-4}$ and weight decay 0.01, mixed-precision training, and 16-frame clips at $224\times 224$. Each client optimizes masked reconstruction on its 34-clip federated train shard ($|D_i|=34$, $q_i=32/34$ under batch size 32).

\textbf{Privacy Configurations.}
We compare a baseline without DP or SA with the DP+SA setting (client-side DP-SGD at $\varepsilon \in \{1,5,10\}$, $\delta=10^{-5}$, plus simulated pairwise masked aggregation with all 40 clients each round). The no-DP/no-SA configuration provides the reference for the privacy-mechanism ablation. Sample-level DP hyperparameters are summarized in Table~\ref{tab:dp_accounting}.

\begin{table}[t]
\centering
\caption{DP-SGD accounting settings on RWF-2000 federated MAE ($|D_i|=34$, $|B|=32$, $E=4$, $R=200$, $\delta=10^{-5}$, $C=1.0$). $\sigma_i$ is auto-calibrated by Opacus from $q_i=\min(1,|B|/|D_i|)$, yielding RDP-accountant estimates under the Poisson-sampling approximation used here.}
\label{tab:dp_accounting}
\small
\setlength{\tabcolsep}{5pt}
\begin{tabular}{@{}lcccc@{}}
\toprule
\textbf{Target $\varepsilon$} & $\sigma_i$ & $q_i$ & $T_i$ & \textbf{RDP estimate} \\
\midrule
1 & 152.5 & $32/34$ & 1600 & $\varepsilon\approx1$ \\
5 & 35.9 & $32/34$ & 1600 & $\varepsilon\approx5$ \\
10 & 19.9 & $32/34$ & 1600 & $\varepsilon\approx10$ \\
\bottomrule
\end{tabular}
\end{table}

\textbf{Evaluation Protocol.}
After federated MAE adaptation, we conduct centralized linear probing with a frozen encoder and linear classifier for 20 epochs, followed by small-head evaluation with a frozen encoder and an MLP containing two hidden layers plus the output layer for 20 epochs. RWF-2000 labels used in both evaluations remain excluded from federation. Table~\ref{tab:results} reports small-head test accuracy, macro-F1, and ROC-AUC. We report linear-probe results separately to characterize the representations before small-head adaptation. All settings use the same VideoMAE architecture and downstream protocol. At $\varepsilon=10$, macro precision, recall, and F1 are 72.25\%, 66.00\%, and 63.43\%.

\textbf{Comparison Protocols.}
We include a \emph{local-only} baseline (mean per-client validation accuracy over 40 clients, each training on 40 local clips and evaluating on five held-out validation clips, with no collaboration), supervised \emph{FedAvg}~\cite{mcmahan2017fedavg} with all 86.2M task parameters trainable (validation-tuned best checkpoint), and \emph{FedProx}~\cite{li2020fedprox} ($\mu=0.01$, 200 rounds, test best) on the same 40-client partition (34 train and 6 validation clips per client). These supervised FL methods provide full-model reference points on the same client partition. Table~\ref{tab:comparison} identifies the learning protocol and evaluation split for each result.

\begin{table}[b]
\centering
\caption{Centralized small-head evaluation on the RWF-2000 test set after federated MAE adaptation. Bold indicates the best value per column.}
\label{tab:results}
\begin{tabular*}{\textwidth}{@{\extracolsep{\fill}}lccc@{}}
\toprule
\textbf{Privacy Setting} & \textbf{Accuracy} & \textbf{Macro-F1} & \textbf{ROC-AUC} \\
\midrule
Baseline (No-DP and No-SA) & \textbf{77.25\%} & \textbf{77.24\%} & \textbf{85.42\%} \\
$\varepsilon=1$ & 65.25\% & 65.04\% & 71.15\% \\
$\varepsilon=5$ & 65.25\% & 65.04\% & 71.15\% \\
$\varepsilon=10$ & 66.00\% & 63.43\% & 71.89\% \\
\bottomrule
\end{tabular*}
\end{table}

\begin{table}[!ht]
\centering
\caption{System comparison across heterogeneous datasets and protocols. RWF-2000 rows report validation or test accuracy as marked. Payload ratios compare the 5.52M federated state with the 156.37M instantiated state ($28.3\times$) and 86.2M FedAvg/FedProx updates ($15.6\times$).}
\label{tab:comparison}
\resizebox{\textwidth}{!}{%
\begin{tabular}{l|ccc|cc|ccc|cc}
\toprule
\textbf{Method} & \textbf{FL} & \textbf{DP} & \textbf{SA} & \textbf{SSL} & \textbf{PEFT} & \textbf{Dataset} & \textbf{Clients} & \textbf{Params} & \textbf{Payload} & \textbf{Accuracy} \\
\midrule
\multicolumn{11}{l}{\textit{Federated Video Learning Baselines}} \\
FedVSSL~\cite{rehman2022fedvssl} & \cmark & \xmark & \xmark & \cmark & \xmark & UCF-101 & 10 & 100\% & $1\times$ & --- \\
FedMAE~\cite{yan2023fedmae} & \cmark & \xmark & \xmark & \cmark & Partial & Kinetics & 20 & $\sim 10\%$ & $20$--$30\times$ & --- \\
\midrule
\multicolumn{11}{l}{\textit{RWF-2000 Baseline Results}} \\
Local-only VideoMAE & \xmark & \xmark & \xmark & --- & --- & RWF-2000 & 40 & --- & --- & 63.50\% (5-clip client val) \\
FedAvg full model~\cite{mcmahan2017fedavg} & \cmark & \xmark & \xmark & --- & \xmark & RWF-2000 & 40 & 100\% & $1\times$ & 88.25\% (val) \\
FedProx full model~\cite{li2020fedprox} & \cmark & \xmark & \xmark & --- & \xmark & RWF-2000 & 40 & 100\% & $1\times$ & 85.75\% (test) \\
\midrule
\multicolumn{11}{l}{\textit{Privacy-Preserving Federated Video (Ours)}} \\
\textbf{Ours (No-DP and No-SA)} & \cmark & \xmark & \xmark & \cmark & \cmark & RWF-2000 & 40 & \textbf{3.5\%} & $\bm{28.3\times}$ & 77.25\% (test) \\
\textbf{Ours (DP+SA, $1 \leq \varepsilon \leq 10$)} & \cmark & \cmark & \cmark & \cmark & \cmark & RWF-2000 & 40 & \textbf{3.5\%} & $\bm{28.3\times}$ & 65.25--66.00\% (test) \\
\bottomrule
\end{tabular}%
}
\end{table}

\begin{figure}[!t]
\centering
\includegraphics[width=\textwidth]{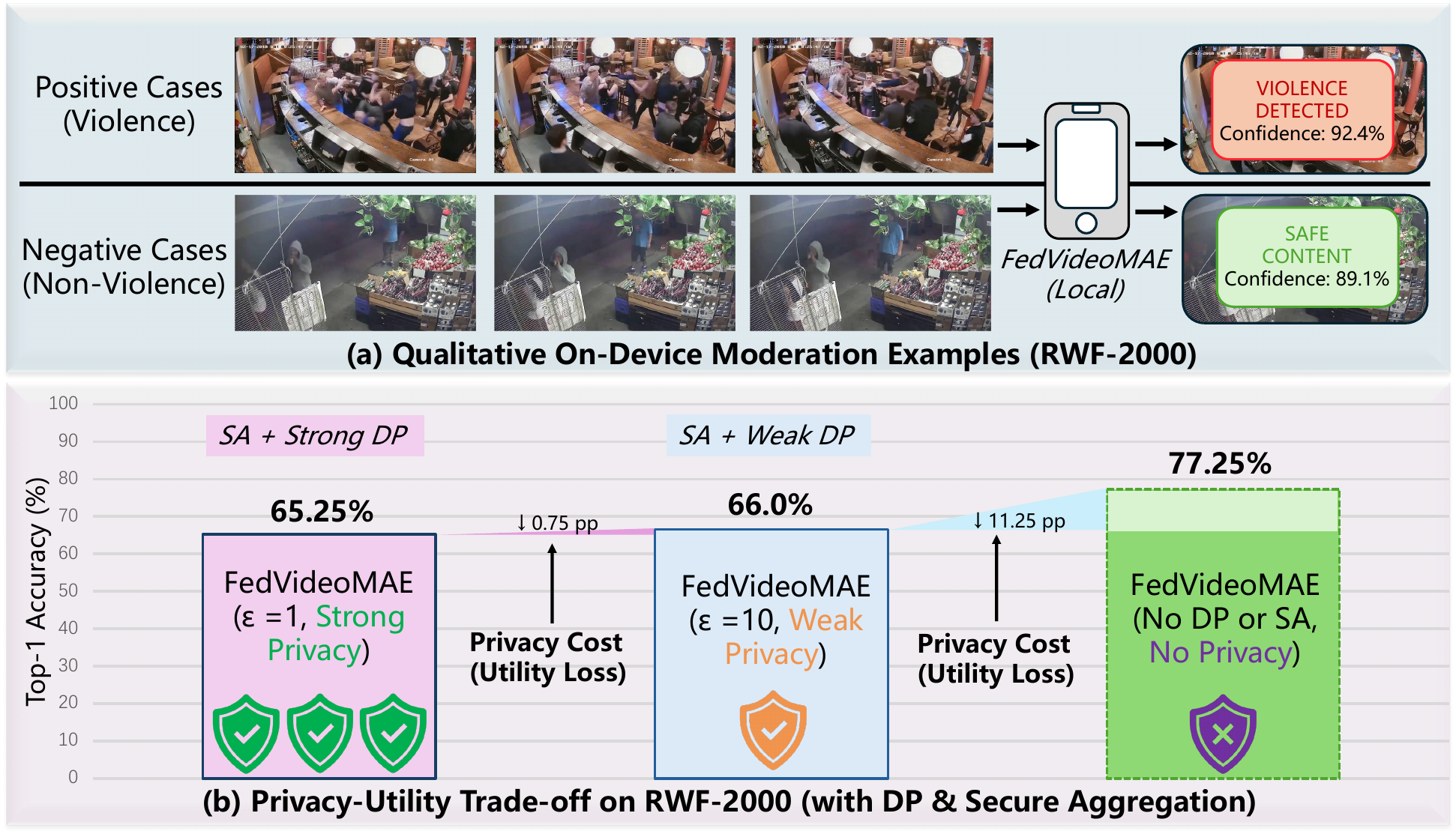}
\caption{Two representative moderation examples and the corresponding privacy--utility summary on RWF-2000. Accuracy differences are expressed in percentage points (pp).}
\label{fig:visualized_results}
\vspace{-0.8em}
\end{figure}

\subsection{Comparative Results}

\textbf{Overall Performance.}
Table~\ref{tab:results} summarizes \emph{centralized small-head} test performance after federated MAE adaptation. The baseline without DP or SA reaches 77.25\% accuracy and 85.42\% ROC-AUC. Under DP+SA, accuracy is 65.25\% at both $\varepsilon=1$ and $\varepsilon=5$, and 66.00\% at $\varepsilon=10$, while ROC-AUC remains above 71\% in all DP settings. At $\varepsilon=10$, macro precision, recall, and F1 are 72.25\%, 66.00\%, and 63.43\%, respectively. Figure~\ref{fig:visualized_results} presents representative moderation examples together with a visual summary of the reported accuracy values.

\textbf{Comparison with Baselines.}
Table~\ref{tab:comparison} places FedVideoMAE alongside prior federated video systems and RWF-2000 references. FedAvg and FedProx use supervised full-model training with 86.2M task parameters, while FedVideoMAE performs federated MAE adaptation followed by centralized head evaluation, with the evaluation split and communication payload reported explicitly for each method. Under these protocols, the no-DP/no-SA configuration reaches 77.25\% small-head test accuracy, while local-only VideoMAE reports 63.50\% mean per-client validation accuracy.

\textbf{Privacy--Utility Trends.}
Relaxing $\varepsilon$ from 1 to 5 leaves small-head accuracy at 65.25\% in Table~\ref{tab:results}, although the calibrated noise multiplier decreases from 152.5 to 35.9. The two settings therefore occupy the same observed accuracy regime under the 34-clip client shards, batch size of 32, and compact adapter configuration, as summarized with the other reported accuracy values in Fig.~\ref{fig:visualized_results}.

\textbf{Auxiliary Transfer.}
Table~\ref{tab:aux_results} reports $\varepsilon=1$ transfer with the same evaluation protocol as RWF-2000. On near-domain RLVS, linear probing reaches 71.50\% accuracy and small-head evaluation improves to 72.00\%, so much of the discriminative structure is already linearly accessible in the frozen representation. On binary UCF-Crime, accuracy is lower for both the linear probe and small-head evaluation (52.50\% and 57.73\%, respectively), but average precision remains relatively high (74.41\% vs.\ 58.99\% ROC-AUC). The scores still carry ranking information under domain shift, even when fixed thresholds degrade.

\begin{table}[!t]
\centering
\caption{$\varepsilon=1$ FedVideoMAE transfer on auxiliary datasets. Linear probe reports accuracy, macro-F1, ROC-AUC, and AP, while small-head reports accuracy, macro-F1, precision, and recall.}
\label{tab:aux_results}
\small
\setlength{\tabcolsep}{4pt}
\begin{tabular}{@{}l ccc ccc@{}}
\toprule
\textbf{Dataset} & \multicolumn{3}{c}{\textbf{Linear probe}} & \multicolumn{3}{c}{\textbf{Small-head}} \\
\cmidrule(lr){2-4}\cmidrule(lr){5-7}
& \textbf{Acc.} & \textbf{F1} & \textbf{AUC, AP} & \textbf{Acc.} & \textbf{F1} & \textbf{Prec., Rec.} \\
\midrule
RLVS & 71.50\% & 71.44\% & 79.33\%, 79.52\% & 72.00\% & 70.71\% & 76.71\%, 72.00\% \\
UCF-Crime & 52.50\% & 52.50\% & 58.99\%, 74.41\% & 57.73\% & 56.13\% & 56.74\%, 57.47\% \\
\bottomrule
\end{tabular}
\end{table}

\subsection{Communication Efficiency and Effective-SNR Analysis}

Each round uploads only LoRA and prompt deltas, while base-model weights remain device-resident and no raw clips or intermediate transformer states are transmitted. The instantiated pretraining model contains 156,371,328 parameters, of which 5,518,848 are federated trainable parameters. Their ratio gives the $28.3\times$ model-state payload reduction reported in \mbox{Table~\ref{tab:comparison}}. The same 5.52M payload is approximately $15.6\times$ smaller than the 86.2M-parameter \mbox{FedAvg} and \mbox{FedProx} updates. These comparisons count exchanged parameter states only and exclude mask-generation and aggregation metadata.

The effective-SNR heuristic relates local data scale, DP noise, trainable-state size, and client participation. On RWF-2000, the calibrated noise multiplier decreases by approximately $4.25\times$ from $\varepsilon=1$ to $\varepsilon=5$, while both settings yield 65.25\% small-head accuracy. This observation places the reported privacy--utility results in the context of the local-data and adapter-scale configuration.

\subsection{Ablation on Privacy Mechanisms}

We compare the baseline, SA-only, DP-only, and DP+SA configurations under the same RWF-2000 40-client schedule and centralized small-head evaluation, using $\varepsilon=5$ for both DP settings. The baseline matches Table~\ref{tab:results} at 77.25\% accuracy and 77.24\% macro-F1. SA-only is 3.00 accuracy points and 3.15 macro-F1 points below the baseline, whereas DP-only is lower by 11.25 and 11.29 points, respectively. DP+SA reaches 65.25\% accuracy and 65.04\% macro-F1, which are 0.75 and 0.91 points below DP-only and 12.00 and 12.20 points below baseline.

\begin{table}[!t]
\centering
\caption{Ablation study of DP-SGD and pairwise masked aggregation on RWF-2000.}
\label{tab:ablation}
\begin{tabular*}{\textwidth}{@{\extracolsep{\fill}}lcccc@{}}
\toprule
\textbf{Method} & \textbf{DP} & \textbf{SA} & \textbf{Acc.} & \textbf{Macro-F1} \\
\midrule
Baseline (No-DP and No-SA) & \xmark & \xmark & \textbf{77.25} & \textbf{77.24} \\
SA-only (No-DP) & \xmark & \cmark & 74.25 & 74.09 \\
DP-only ($\varepsilon=5$) & $\varepsilon=5$ & \xmark & 66.00 & 65.95 \\
DP+SA ($\varepsilon=5$) & $\varepsilon=5$ & \cmark & 65.25 & 65.04 \\
\bottomrule
\end{tabular*}
\end{table}

Relative to DP-only, DP+SA changes accuracy by 0.75 percentage points and macro-F1 by 0.91 points. Enabling DP-SGD produces the larger difference from the no-DP/no-SA reference: 11.25 points in accuracy for DP-only and 12.00 points for DP+SA. When all participating clients contribute, pairwise masks cancel during aggregation, yielding the averaged adapter parameters.

Across the four configurations, the larger performance change appears when DP-SGD is enabled, whereas adding pairwise masking to the same DP setting changes accuracy by 0.75 points and macro-F1 by 0.91 points. This pattern associates the dominant observed utility reduction with the configurations using DP-SGD under the evaluated schedule.

\section{Discussion and Limitations}
\label{sec:discussion}

\textbf{Privacy Scope.}
FedVideoMAE separates representation learning from supervised moderation evaluation. During federation, each client processes its own clips, optimizes LoRA and prompt parameters through masked reconstruction, and retains raw video, intermediate features, and labels locally. The server receives adapter states after local DP-SGD and pairwise masking, aggregates them with FedAvg, and broadcasts the resulting global state for the next round. During evaluation, the adapted encoder is frozen and a trusted labeled holdout supplies the supervision needed for linear probing, small-head training, and threshold selection. This two-stage flow keeps violence labels outside the federated MAE objective while supporting downstream moderation assessment of the learned representation in the centralized stage.

The two privacy mechanisms act at different points in this update path. Under the schedule in Table~\ref{tab:dp_accounting}, DP-SGD clips per-sample adapter gradients and adds Gaussian noise during local optimization, while pairwise masks are applied to the resulting adapter states immediately before FedAvg. The reported $\varepsilon$ values describe samples within each client's shard under the stated RDP-accounting approximation. The ablation then measures the utility changes associated with DP-SGD and pairwise masking under this update sequence.

\textbf{Baseline and Communication Context.}
Table~\ref{tab:comparison} organizes the reference systems by learning objective, trainable parameter scope, evaluation split, and communication payload. FedAvg and FedProx optimize supervised full-model updates, whereas FedVideoMAE federates self-supervised adapter states before centralized head evaluation, and local-only VideoMAE provides the non-collaborative reference on small client datasets. Because these protocols answer different questions, the full-model rows show the accuracy available when all task parameters participate in supervised federation, the local-only row measures learning without collaboration, and the FedVideoMAE rows quantify the result of label-free federated adaptation with a compact trainable state.

The $28.3\times$ ratio compares the 5.52M federated state with the instantiated 156.37M-parameter pretraining model, while the $15.6\times$ ratio compares it with the 86.2M-parameter FedAvg and FedProx updates. Both ratios describe exchanged parameter states and use the same numerator throughout the paper. Read together with the marked validation and test splits, these dimensions position FedVideoMAE between local-only training and communication-heavy full-model FL. Within the evaluated protocols, these comparisons quantify the communication advantage of adapting only LoRA and prompt states while keeping the VideoMAE backbone fixed.

\textbf{Experimental Scope and Practical Implications.}
With approximately 3.5\% federated parameters, FedVideoMAE retains 65--66\% small-head test accuracy under DP+SA, compared with 77.25\% without privacy mechanisms. The identical accuracy at $\varepsilon=1$ and $5$, despite the lower calibrated noise multiplier, places both settings in the same observed utility regime for the 34-clip shards and compact adapter configuration. In this regime, each client contributes nearly its entire local shard to a minibatch, repeats local optimization across 200 communication rounds, and applies DP noise only to the compact trainable subspace. The effective-SNR heuristic collects these interacting factors into a common optimization view for interpreting why changing $\varepsilon$ alone need not produce a proportional accuracy change.

The controlled setup uses 40 full-participation clients and fixed shards, which keeps the accounting schedule and mask-cancellation condition consistent across the privacy ablation. The ablation shows the larger utility change when DP-SGD is enabled, while pairwise masking changes accuracy by less than one point when added to the same DP setting. The auxiliary results provide a complementary transfer view: RLVS remains close to the RWF-2000 violence task, whereas binary UCF-Crime contains longer videos and a broader abnormality label. The difference between their linear-probe and small-head results indicates how the frozen representation responds to both near-domain and shifted evaluation conditions under the same $\varepsilon=1$ protocol.

For an edge moderation deployment, the frozen VideoMAE backbone can remain resident on each device while communication is limited to LoRA and prompt states. This arrangement moves masked reconstruction and privacy processing to the client and reserves labeled classifier training for the centralized evaluation stage. The measured parameter ratios quantify the reduction in exchanged model state, while the reported accuracy, macro-F1, and ROC-AUC describe the corresponding downstream utility. Larger local datasets and alternative PEFT capacities directly test the data and model factors represented in the effective-SNR heuristic, while partial participation and stronger non-IID partitions extend the aggregation and accounting conditions studied here. Repeated trials would further characterize the variability of the sub-one-point differences observed between DP-only and DP+SA.

\section{Conclusion}

We presented FedVideoMAE, which combines frozen VideoMAE base weights with trainable LoRA and prompt adaptation for federated masked video reconstruction. Client training uses sample-level DP-SGD before pairwise masked aggregation and FedAvg, while violence labels are reserved for centralized downstream evaluation. This separation allows clients to learn from local video without placing the supervised moderation objective inside federation. Exchanging 5,518,848 trainable parameters gives a $28.3\times$ model-state payload ratio and is approximately $15.6\times$ smaller than the full-model FedAvg and FedProx references. Small-head test accuracy reaches 77.25\% without DP or SA and 65--66\% under DP+SA, while the RLVS and binary UCF-Crime experiments extend the evaluation across near-domain and shifted video data. The effective-SNR heuristic relates the observed utility regime to local data scale, privacy noise, PEFT capacity, and client participation, and the ablation associates the largest performance change with configurations using DP-SGD. FedVideoMAE therefore provides an implementation-grounded reference for privacy-aware federated video moderation under limited local data and communication budgets.

\begingroup
\makeatletter
\renewcommand\small{\@setfontsize\small\@ixpt{10\p@}}
\makeatother
\bibliographystyle{splncs04}
\bibliography{ref}
\endgroup
\end{document}